\documentclass{article}
\usepackage{ijcai21}

\usepackage{times}
\usepackage{soul}
\usepackage{url}
\usepackage[hidelinks]{hyperref}
\usepackage[utf8]{inputenc}
\usepackage[small]{caption}
\usepackage{graphicx}
\usepackage{amsmath}
\usepackage{color}
\usepackage{amsthm}
\usepackage{booktabs}
\usepackage{algorithm}
\usepackage{algorithmic}
\usepackage{multirow}
\usepackage{adjustbox}
\usepackage{arydshln}
\usepackage{CJKutf8}
\newcommand{\newcite}[1]{\citeauthor{#1}~\shortcite{#1}}

\title{A Survey on Spoken Language Understanding: Recent Advances and New Frontiers}

\author{
Anonymous IJCAI submission
}

\author{
	Libo Qin\and
	Tianbao Xie\and
	Wanxiang Che\thanks{Corresponding Author}\and Ting Liu
	\affiliations
	Research Center for Social Computing and Information Retrieval \\Harbin Institute of Technology, China\\
	\emails
	\{lbqin, tianbaoxie, car, tliu\}@ir.hit.edu.cn
}

\begin{document}

\maketitle

\begin{abstract}
Spoken Language Understanding (SLU) aims to extract the semantics frame of user queries, which is a core component in a task-oriented dialog system.
With the burst of deep neural networks and the evolution of pre-trained language models, the research of SLU has obtained significant breakthroughs.
However, there remains a lack of a comprehensive survey summarizing existing approaches and recent trends, which motivated the work presented in this article.
In this paper, we survey recent advances and new frontiers in SLU.
Specifically, we give a thorough review of this research ﬁeld, covering different aspects including (1) new taxonomy:  we provide a new perspective for SLU filed, including \textit{single model} vs. \textit{joint model}, \textit{implicit joint modeling} vs. \textit{explicit joint modeling} in joint model,  \textit{non pre-trained paradigm} vs. \textit{pre-trained paradigm};
(2) new frontiers: some emerging areas in complex SLU as well as the corresponding challenges;
(3) abundant open-source resources: to help the community, we have collected, organized the related papers, baseline projects and leaderboard on a public website\footnote{\url{https://github.com/yizhen20133868/Awesome-SLU-Survey}} where SLU researchers could directly access to the recent progress.
We hope that this survey can shed a light on future research in SLU field.
\end{abstract}

\section{Introduction}
\label{Introduction}
\begin{figure}[t]
	\centering
	\includegraphics[scale=0.4]{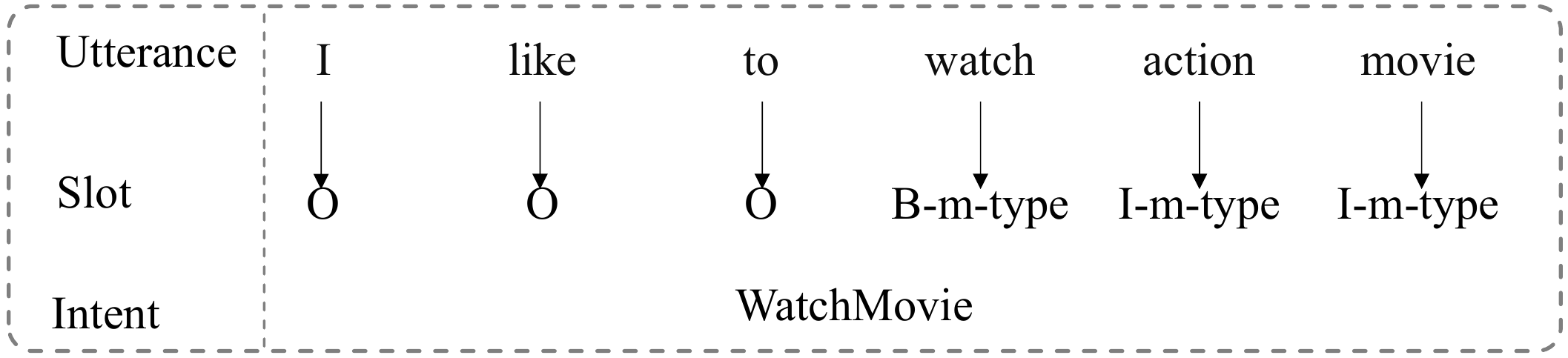}
	\caption{ An example with intent and slot annotation (BIO format).
		m-type denotes movie-type.
	}
	\label{fig:example}
\end{figure}

Spoken Language Understanding (SLU) is a core component in task-oriented dialog system, which
aims to capture the semantics of user queries.
It typically consists of two tasks: intent detection and slot filling \cite{tur2011spoken}. 
Take the utterance \textit{``I like to watch action movie"} in Figure~\ref{fig:example} as an example, 
the outputs include an intent class label (i.e., \texttt{WatchMovie}) and a slot label sequence (i.e., \texttt{O, O, O,  B-movie-type, I-movie-type, I-movie-type}).

Intent detection can be defined as a sentence classification problem. 
In recent years, many neural-network based classification methods such as convolutional neural network (CNN) \cite{xu2013convolutional} and recurrent neural network(RNN) \cite{ravuri2015recurrent} have been investigated.
Slot filling can be formulated as a sequence labeling task and popular sequence labeling methods such as conditional random field (CRF) \cite{raymond2007generative}, RNN-based models \cite{xu2013convolutional} and Long Short-Term Memory Network (LSTM) \cite{ravuri2015recurrent}  have been explored.

\begin{figure}[t]
	\centering
	\includegraphics[width=0.45\textwidth]{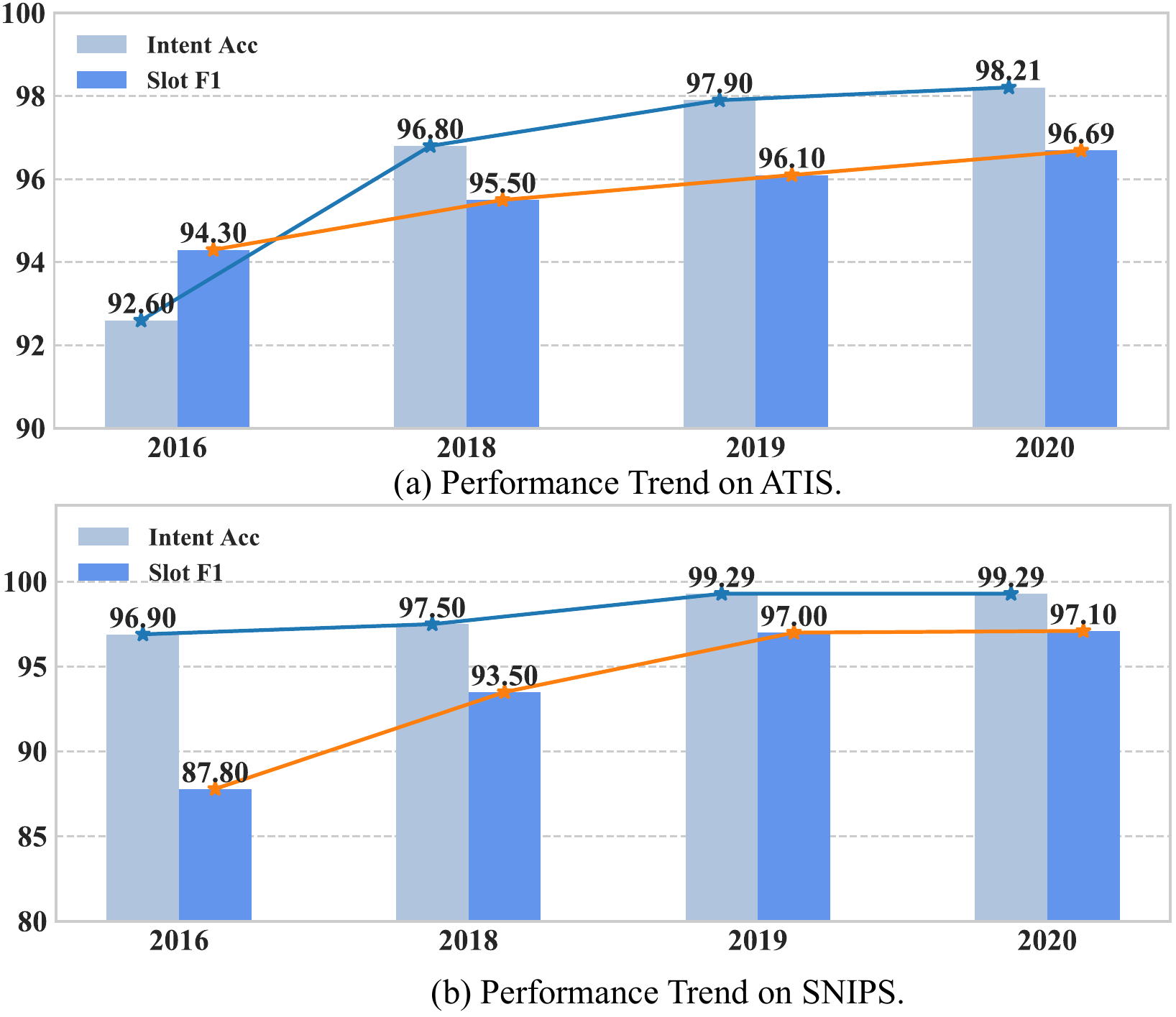}
	\caption{ Recent Performance Trend.
	}
	\label{fig:performanc_atis}
\end{figure}

Traditional approaches consider slot filling and intent detection as two separate tasks, which 
ignore the shared knowledge across the two tasks.
Intuitively, intent detection and slot filling are not independent and highly tied.
For example, if the intent of a user query is \texttt{WatchMovie}, it is more likely to contain the slot movie name rather than the slot music name.
Thus, it's promising to consider the interaction between the two tasks.
To this end, dominant models in the literature adopt joint models for leveraging shared knowledge across the two tasks, such as vanilla multi-task \cite{zhang2016joint}, slot-gated \cite{goo-etal-2018-slot,li-etal-2018-self}, stack-propagation \cite{qin-etal-2019-stack} and bi-directional interaction \cite{e-etal-2019-novel,qin2020cointeractive}.
With the popularity of deep learning and the emergence of pre-trained language models, SLU direction has made significant progress in recent years.
As shown in Figure~\ref{fig:performanc_atis}, in slot filling and intent detection tasks, we clearly observe that performance has even surpassed 97.0\% and 98.0\% on ATIS~\cite{hemphill-etal-1990-atis} while 97\% and 99\% on SNIPS~\cite{coucke2018snips} that are the most wildly used datasets in SLU community. 
This leaves us with a question: \textit{Have we achieved SLU tasks perfectly}?

In this paper, we introduce a survey to answer the above question including: 1) a comprehensive summary of recent progress in SLU field; 2) research challenges and frontiers for complex SLU tasks are concluded.
Our survey observes that mainstream work remains the simple setting: single domain and single turn, which is still far from satisfying the requirements of some complex applications.

In summary, the contributions of this survey can be concluded as follows:
\begin{itemize}
	\item \textit{New taxonomy}. We propose a taxonomy of SLU field, which categorizes existing approaches from three different perspectives: 1) \textit{single model} vs. \textit{joint model}; 2) \textit{implicit joint model} vs. \textit{explicit joint model} in joint model; 3) \textit{non pre-trained models} vs. \textit{pre-trained models}.
	
	\item \textit{Abundant resources}. We collect abundant resources on SLU including open-source implementations, corpora, and paper lists.
	To our knowledge, this is the first effort to collect open-source resources for SLU community.
	
	\item \textit{New Frontiers}. We discuss and analyze the limitations of existing SLU. Also, we suggest some new frontiers and discuss the challenges.
\end{itemize}

We hope this survey will help researchers to understand the latest progress, challenges and frontiers in SLU field.

\begin{figure*}[t]
	\centering
	\includegraphics[width=\linewidth]{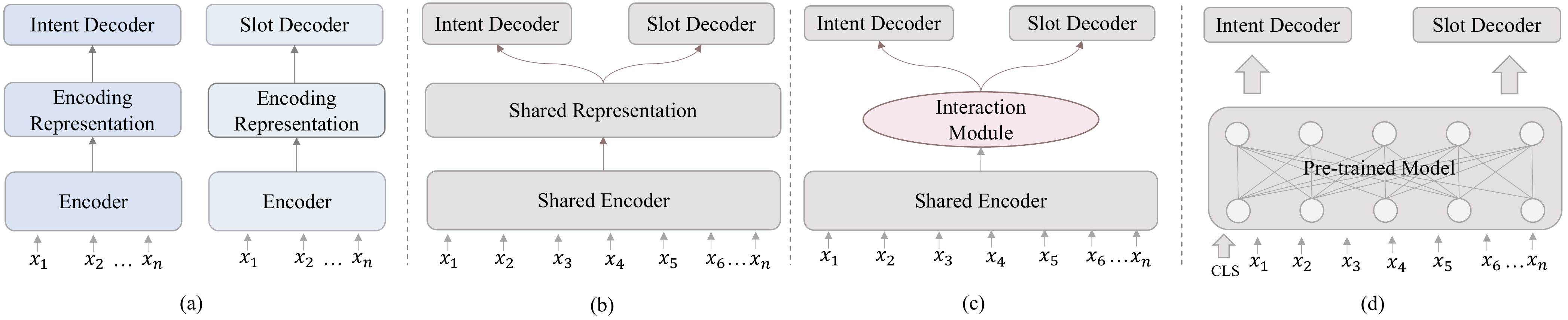}
	\caption{
		(a) Single models.	(b) Implicit Joint Modeling. (c) Explicit Joint Modeling. (d) Pre-trained Model Paradigm. \\
	}
	\label{fig:contrast}
\end{figure*}
The rest of the survey is organized as follows. Section~\ref{background} outlines the background of SLU. Section~\ref{taxonomy} gives a brief overview proposed taxonomy of SLU. Section~\ref{complex-setting} discusses the new frontiers and their challenges. Section~\ref{related-work} gives the related survey on SLU. Section~\ref{conclusion} summarizes the paper.

\section{Background}
\label{background}

\begin{table}[t]
	\centering
	\setlength{\tabcolsep}{1mm}{
		\begin{tabular}{lcc}
			\toprule
			Model  & Intent Acc & Slot F1 \\
			\midrule
			Bi-Jordan RNN \cite{mesnil2013investigation} & - & 93.98      \\
			RNN \cite{yao2013recurrent} & - & 94.11      \\
			Hybrid RNN \cite{mesnil2014using} & - & 95.06      \\
			LSTM \cite{yao2014spoken} & - & 95.08      \\
			R-CRF \cite{yao2014recurrent} & - & 96.65      \\
			RNN \cite{ravuri2015recurrent} & 97.55 & -      \\
			LSTM \cite{ravuri2015recurrent} & 98.06 & -      \\
			RNN SOP \cite{liu2015recurrent} & - & 94.89      \\
			5xR-biRNN \cite{Vu2016bi-directional} & - & 95.56      \\
			Encoder-labeler \cite{kurata-etal-2016-leveraging} & - & 95.66      \\
			\bottomrule
	\end{tabular}}
	\caption{Single model performance on intent detection and slot filling on ATIS. Acc denotes the accuracy metric.}
	\label{tab:single model performance}
\end{table}
In this section, we describe the deﬁnition for slot filling, intent detection and joint model, and then we give a brief description of the wildly used datasets and evaluation metrics.
\subsection{Definition}
\paragraph{Intent Detection:}
Given input utterance $X$ = $(x_{1},\dots,x_{n})$ ($n$ denotes the length of $X$), intent detection (ID) can be considered as a sentence classification task to decide the intent label $o^{I}$, which is formulated as:
\begin{equation}
	o^{I} =  \texttt{Intent-Detection} (X).
\end{equation}
\paragraph{Slot Filling:}
 Slot filling (SF) can be seen as a sequence labeling task to produce a sequence slots $o^{S}$ = $(o^{S}_{1},\dots,o^{S}_{n})$, which can be written as:
\begin{equation}
o^{S} = \texttt{Slot-Filling} (X).
\end{equation}
\paragraph{Joint Model:}
Joint model denotes that a joint model predicts the slots sequence and intent simultaneously, which has the advantage of capturing shared knowledge across related tasks, using:
\begin{equation}
(o^{I}, o^S)  =  \texttt{Joint-Model} (X).
\end{equation}

\subsection{Dataset}
The most wildly used datasets are ATIS \cite{hemphill-etal-1990-atis} and SNIPS \cite{coucke2018snips}.
In the following, we will give a detailed description.
\paragraph{ATIS:}
ATIS dataset contains audio recordings of ﬂight, reservations.
There are 4,478 utterances for training, 500 utterances for validation and 500 utterances for testing.
120 slot labels and 21 intent types are included in ATIS training data.
\paragraph{SNIPS:}
SNIPS is the custom-intent-engines collected by Snips \cite{coucke2018snips}.
There are 13,084 utterances for training, 700 utterances for validation and 700 utterances for testing.
There are 72 slot labels and 7 intent types. 
\subsection{Evaluation Metrics}
The most wildly used evaluation metrics for SLU are F1 scores, intent accuracy and overall accuracy.
\begin{itemize}
	\item \textbf{F1 scores:} F1 scores are adopted to evaluate the performance of slot filling, which is the harmonic mean score between precision and recall.
	A slot prediction is considered correct when an exact match is found  \cite{tjong-kim-sang-de-meulder-2003-introduction}.
	\item \textbf{Intent Accuracy:} Intent Accuracy is used for evaluating the performance of intent detection, calculating the ratio of sentences for which intent is predicted correctly.
	\item \textbf{Overall  Accuracy:} Overall accuracy is adopted for calculating the ratio of sentences for which both intent and slot are predicted correctly in a sentence \cite{goo-etal-2018-slot}.
	This metric considers intent detection and slot filling simultaneously.
\end{itemize}

\section{Taxonomy}
\label{taxonomy}

In this section, we describe the taxonomy in SLU, including \textit{single models} ($\S \ref{single-model}$), \textit{joint models} ($\S \ref{joint-model}$) and \textit{pre-trained paradigm} ($\S \ref{pre-trained-model}$), which is shown in Figure~\ref{fig:contrast}.
\subsection{Single Model}
\label{single-model}

Single models train each task \textit{separately} for intent detection and slot filling, which is shown in Figure~\ref{fig:contrast}(a).
\paragraph{Intent Detection:}
Many sentence classification methods have been investigated for intent detection. \newcite{xu2013convolutional} utilized Convolutional Neural Network (CNN)~\cite{lecun1998gradient} to extract 5-gram features and apply max-pooling to obtain word representations.
\newcite{ravuri2015recurrent} successfully applied RNN and Long Short-Term Memory Network(LSTM)~\cite{hochreiter1997long} to the ID task, which indicates the sequential features are beneficial to intent detection task.

\paragraph{Slot Filling:}
Popular neural approaches for slot filling including CRF, Recurrent neural network (RNN) and RNN-based models.
\newcite{yao2013recurrent} adopted RNN Language Models ({RNN-LMs}) to predict slot labels rather than words. In addition, RNN-LMs explored future words, named entities, syntactic features and word-class information.
\newcite{mesnil2013investigation} investigated several RNN architectures including Elman RNN, Jordan RNN and its bi-directional version for SLU.
\newcite{yao2014spoken} proposed an LSTM framework for the slot filling task.
\newcite{mesnil2014using} applied Viterbi encodings and recurrent CRFs to eliminate the label bias problem.
\newcite{yao2014recurrent} proposed R-CRF to tackle label bias.
\newcite{liu2015recurrent} proposed to model slot label dependencies using a sampling approach, by feeding sampled output labels (true or predicted) back to the sequence state.
\newcite{Vu2016bi-directional} utilized ranking loss function on bi-RNN model in SF, further enhancing performance in ATIS dataset.
\newcite{kurata-etal-2016-leveraging} leveraged sentence-level information from encoder to improve performance for SF task.

The wildly used dataset for evaluating single models is ATIS.
Table \ref{tab:single model performance} summarizes the performance of single model on intent detection and slot filling.

\paragraph{Highlight:}
There is no interaction between intent detection and slot filling in single models due to the separate training, leading to shared knowledge leakage across two tasks.

\subsection{Joint Model}
\label{joint-model}

Considering the close correlation between intent detection and slot filling, dominant work in the literature adopts joint model to leverage the shared knowledge across tasks.
Existing joint work can be classified into two main categories: \textit{implicit joint modeling} and \textit{explicit joint modeling}.
\begin{table*}[h]
	\centering
	\begin{adjustbox}{width=\textwidth}
		\begin{tabular}{l||ccc||ccc}
			\hline
			& \multicolumn{3}{c||}{ATIS} & \multicolumn{3}{c}{SNIPS} \\
			\hline \hline
			Model&
			Intent Acc & Slot F1 & Overall Acc & 
			Intent Acc & Slot F1 & Overall Acc 
			\\ 
			\hline \hline
			\multicolumn{7}{c}{\textit{Implicit Joint Modeling}} \\
			\hline
	Joint ID and SF \cite{zhang2016joint}
& 98.32 & 96.89 & - & - & - & -\\
	Attention BiRNN \cite{liu2016attention}
& 91.10 & 94.20 & 78.90 & 96.70 & 87.80 & 74.10  \\
			Joint SLU-LM \cite{liu-lane-2016-joint}
			& 98.43 & 94.47 & - & - & - & -\\
			Joint Seq. \cite{hakkani2016multi}
			& 92.60 & 94.30 & 80.70 & 96.90 & 87.30 & 73.20 \\

			\hline
			\multicolumn{7}{c}{\textit{Explicit Joint Modeling}} \\
			\hline
			Slot-Gated \cite{goo-etal-2018-slot}
			& 93.60 & 94.80 & 82.20 & 97.00 & 88.80 & 75.50 \\
			Self-Atten. Model \cite{li-etal-2018-self}
			& 96.80 & 95.10 & 82.20 & 97.50 & 90.00 & 81.00\\
			Bi-model \cite{wang-etal-2018-bi}
			& 96.40 & 95.50 & 85.70 & 97.20 & 93.50 & 83.80\\
			SF-ID Network \cite{e-etal-2019-novel}
			& 97.09 & 95.80 & 86.90 & 97.29 & 92.23 & 80.43\\
			Capsule-NLU \cite{zhang-etal-2019-joint}
			& 95.00 & 95.20 & 83.40 & 97.30 & 91.80 & 80.90\\
			CM-Net \cite{liu-etal-2019-cm}
			& 96.10 & 95.60 & 85.30 & 98.00 & 93.40 & 84.10\\
			Stack-Propgation \cite{qin-etal-2019-stack}
			& 96.90 & 95.90 & 86.50 & 98.00 & 94.20 & 86.90\\
			Graph LSTM \cite{zhang2020graph}
			& 97.20 & 95.91 & 87.57 & 98.29 & 95.30 & 89.71 \\
			Co-Interactive transformer \cite{qin2020cointeractive}
			& 97.70 & 95.90 & 87.40 & 98.80 & 95.90 & 90.30\\
			\hline \hline
			\multicolumn{7}{c}{\textit{Pre-trained Models}} \\
			\hline
			BERT-Joint \cite{castellucci2019multilingual}
			& 97.80 & 95.70 & 88.20 & 99.00 & 96.20 & 91.60\\
			Joint BERT +CRF \cite{chen2019bert}
			& 97.90 & 96.00 & 88.60 & 98.40 & 96.70 & 92.60\\
			Stack-Propgation +BERT \cite{qin-etal-2019-stack}
			& 97.50 & 96.10 & 88.60 & 99.00 & 97.00 & 92.90\\
			Co-Interactive transformer +BERT \cite{qin2020cointeractive}
			& 98.00 & 96.10 & 88.80 & 98.80 & 97.10 & 93.10 \\
			\hline 
		\end{tabular}
	\end{adjustbox}
	\caption{Joint model performance on intent detection and slot ﬁlling. Acc denotes the accuracy metric. We adopted reported results from published literature \protect\cite{goo-etal-2018-slot} and \protect\cite{qin2020cointeractive}.} 
	\label{tab:joint model performance}
\end{table*}
\paragraph{\textit{Implicit Joint Modeling:}}
Implicit joint modeling denotes that model only adopts a shared encoder to capture shared features, without any explicit interaction, which is illustrated in Figure~\ref{fig:contrast}(b).
\newcite{zhang2016joint} introduced a shared RNNs ({Joint ID and SF}) to learn the correlation between intent and slots.
\newcite{liu2016attention} introduced a shared encoder decoder framework with attention-mechanism ({Attention BiRNN}) for intent detection and slot filling.
\newcite{liu-lane-2016-joint} used a shared RNN to jointly perform SF, ID and language modeling ({Joint SLU-LM}), aiming to improve the ability of online prediction.
\newcite{hakkani2016multi} proposed a shared RNN-LSTM architecture ({Joint Seq}) for joint modeling.
\paragraph{Highlight:} Though \textit{implicit joint modeling} is a direct method to incorporate the shared knowledge, it does not model the interaction explicitly, resulting in low interpretability and low performance.

\paragraph{\textit{Explicit Joint modeling:}}
In recent years, more and more work has been proposed to explicitly model the interaction between intent detection and slot filling with an explicit interaction module, which is shown in Figure~\ref{fig:contrast}(c).
This explicit modeling mode has the advantages of explicitly controlling process of interaction.
The existing \textit{explicit joint modeling} methods can be categorized into two types: \textit{single flow interaction} and \textit{bidirectional flow interaction}.
\begin{itemize}
	\item \textit{\textbf{Single Flow Interaction:}}
	Recent work on \textit{single flow interaction} mainly considered the single information ﬂow from intent to slot. \newcite{goo-etal-2018-slot} proposed a slot-gated joint model ({Slot-Gated}), which allows the slot filling be can be conditioned on the learned intent.
	\newcite{li-etal-2018-self} proposed a novel self-attentive model ({Self-Atten. Model}) with the intent augmented gate mechanism to guide the slot filling.
	\newcite{qin-etal-2019-stack} proposed a stack-propagation model to directly use intent detection results to guide slot filling and used the token-level intent detection to alleviate the error propagation, further enhancing the performance.
	
	\item \textit{\textbf{Bidirectional Flow Interaction:}}
	Bidirectional flow interaction work means that model considered the cross-impact between intent detection and slot ﬁlling. As exploration, \newcite{wang-etal-2018-bi} proposed a Bi-Model architecture to consider the cross-impact across SF and ID by using two correlated bidirectional LSTMs.
	\newcite{e-etal-2019-novel} proposed a novel SF-ID network that provides a bi-directional interrelated mechanism for SF and ID tasks, considering the influence of SF-to-ID and ID-to-SF.
	\newcite{zhang-etal-2019-joint} introduced a dynamic routing capsule network ({Capsule-NLU}) to incorporate hierarchical and interrelated relationships among two tasks.
	\newcite{liu-etal-2019-cm} proposed a novel collaborative memory network ({CM-Net}) for jointly modeling SF and ID.
	\newcite{zhang2020graph} do exploration in introducing graph LSTM to SLU, achieving the promising performance.
	\newcite{qin2020cointeractive} proposed a co-interactive transformer to consider the cross-impact by building a bidirectional connection between the two related tasks. 
	
\end{itemize}
The wildly used datasets for evaluating joint models are ATIS and SNIPS.
	Table \ref{tab:joint model performance} concludes the performance.

\paragraph{Highlight:} Compared with \textit{implicit joint modeling} method, \textit{explicit joint modeling} has the following advantages. First, a simple multi-task framework just implicitly considers mutual connection between two tasks by sharing latent representations, which cannot achieve desirable results. In contrast, \textit{explicit joint modeling} can enable model to fully capture the shared knowledge across tasks, which promotes the performance on two tasks.
Second, explicitly controlling knowledge transfer for two tasks can help to improve interpretability where impact between SF and ID can be analyzed easily.

\subsection{Pre-trained Paradigm}
\label{pre-trained-model}
Recently, Pre-trained Language Models (PLMs) achieve surprising results across various NLP tasks \cite{wang2020from}.
Some BERT-based \cite{devlin-etal-2019-bert} pre-trained work has been explored in SLU direction where a shared BERT is considered as the encoder to extract contextual representations.
In BERT-based models, each utterance starts with \texttt{[CLS]} and ends with \texttt{[SEP]}, where \texttt{[CLS]} is the special symbol for representing the whole sequence, and \texttt{[SEP]} is the special symbol to separate non-consecutive token sequences.
Further, the representation of the special token \texttt{[CLS]} is used for intent detection while other token representations are adopted for slot filling, which is shown in Figure~\ref{fig:contrast}(d).

More specifically, \newcite{chen2019bert} explored BERT for SLU 
where BERT is used to extract shared contextual embedding for intent detection and slot filling, which obtains a significant improvement compared with other non pre-trained models.
\newcite{castellucci2019multilingual} used the simlilar architecture ({BERT-Joint}) for jointly modeling intent detection and slot filing.
\newcite{qin-etal-2019-stack} used pre-trained embedding encoder to replace its attention encoder ({Stack-Propgation +BERT}), further boosting model's performance. 
\newcite{qin2020cointeractive} also explored BERT for SLU ({Co-Interactive transformer +BERT}), which achieves state-of-the-art performance.
Table~\ref{tab:joint model performance} shows the results of pre-trained models.

\paragraph{Highlight:} Pre-trained models can provide rich semantic features, which can help to improve the performance on SLU tasks.
This observation is consistent with pre-trained models for other NLP applications.

\section{New Frontiers and Challenges}
\label{complex-setting}
Section~\ref{taxonomy} has discussed the traditional SLU setting that mainly focuses on the single-domain or single-turn setting, which limits its application and may be not enough for complex needs in the real-world scenario.
In the following, we will discuss new frontiers in a complex setting and their challenges, including \textit{contextual SLU} ($\S \ref{sec:contextual}$), \textit{multi-intent SLU} ($\S \ref{sec:multi-intent}$), \textit{Chinese SLU} ($\S \ref{sec:chinese}$), \textit{cross-domain SLU} ($\S \ref{sec:cross-domain}$), \textit{cross-lingual SLU} ($\S \ref{sec:cross-lingual}$) and \textit{low-resource SLU} ($\S \ref{sec:low-resource}$).
\subsection{Contextual SLU}\label{sec:contextual}
Naturally, completing a task usually necessitates multiple turns of back-and-forth conversations between the user and the system, 
 which requires model to consider the contextual SLU.
 Unlike the single turn SLU, contextual SLU faces unique ambiguity challenge, since the user and the system may refer to entities introduced in prior dialogue turns, introducing ambiguity, which requires model to incorporate the contextual information for alleviating ambiguity.

To this end, \newcite{chen2016end-to-end} proposed a memory network to incorporate dialogue history information, showing that their model outperforms models without context. 
\newcite{bapna-etal-2017-sequential} proposed a sequential dialogue encoder network, which allows encoding context from the dialogue history in chronological order.  
\newcite{su-etal-2018-time} designed and investigated various time-decay attention functions based on an end-to-end contextual language understanding model. 
\newcite{9330801} proposed an adaptive fusion layer to dynamically consider the different and relevant contextual information for guiding the slot filling, achieving a fine-grained contextual information transfer.

 The main challenges for contextual SLU are as follows:
 \begin{itemize}
 	\item \textbf{Contextual Information Integration:} 
 	Correctly differentiating relevance between different dialog histories with the current utterance and effectively incorporating contextual information into contextual SLU is a core challenge.
 	
 	\item \textbf{Long Distance Obstacles:}
    Since some dialogues have extreme long histories, how to effectively model the long-distance dialog history and filter irrelevant noise is an interesting research topic.
 \end{itemize}
\subsection{Multi-Intent SLU}\label{sec:multi-intent}
Multi-intent SLU means that the system can handle an utterance containing multiple intents and its corresponding slots.
\newcite{gangadharaiah-narayanaswamy-2019-joint} show that 52\% of examples are multi-intent in the amazon internal dataset, which indicates that multi-intent setting is more practical in the real-world scenario.

To this end, \newcite{gangadharaiah-narayanaswamy-2019-joint} explored a multi-task framework to jointly perform multi-intent classification and slot filling. 
\newcite{qin-etal-2020-agif} proposed an adaptive graph-interactive framework to model the interaction between multiple intents and slot at each token.
 
 There are several possible reasons for the slow progress of multi-intent SLU:
  \begin{itemize}
 	\item \textbf{Interaction between Multiple Intents and Slots:}
    Unlike the single intent SLU, how to effectively incorporate multiple intents information to lead the slot prediction is a unique challenge in multi-intent SLU.
 	
 	\item \textbf{Lack of Data:}
    There is no human-annotated data for multi-intent SLU yet, which is another possible reason for the slow progress. \end{itemize}

\subsection{Chinese SLU}\label{sec:chinese}
Chinese SLU means that an SLU model trained on Chinese data and directly be applied to Chinese community.
Compared with SLU in English, Chinese SLU faces a unique challenge since it usually needs word segmentation.

\newcite{liu-etal-2019-cm} contributed a new corpus (CAIS) to the research community.
In addition, they proposed a character-based joint model to perform Chinese SLU.
Though achieving promising performance, one drawback of the character-based SLU model is that explicit word sequence information is not fully exploited, which can be potentially useful.
\newcite{teng2020injecting} proposed a multi-level word adapter to effectively incorporate word information for both \textit{sentence-level} intent detection and \textit{token-level} slot filling.

 The main challenges for Chinese SLU are as follows:
  \begin{itemize}
 	\item \textbf{Word Information Integration:} How to effectively incorporate word information to guide Chinese SLU is a unique challenge.
 	\item \textbf{Multiple Word Segmentation Criteria:} Since there are multiple word segmentation criteria, how to effectively combine multiple word segmentation information for Chinese SLU is non-trivial.
  \end{itemize}

\subsection{Cross Domain SLU}\label{sec:cross-domain}
Though achieving promising performance in single domain setting, the existing SLU models rely on a considerable amount of annotated data, which limits their usefulness for new and extended domains. In practice, it's infeasible to collect rich labeled datasets for each new domain.
Hence, it's promising to consider the cross-domain setting.

Methods in this area can be concluded into two categories: \textit{Implicit domain knowledge transfer} with parameters sharing and \textit{explicit domain knowledge transfer}. 
\textit{Implicit domain knowledge transfer} means that model simply combines multi-domain datasets for training to capture domain features. 
\newcite{hakkani2016multi} proposed a single LSTM model over mixed multi-domain dataset, which can implicitly learn the domain-shared features.
\newcite{kim2017onenet} adopted one network to jointly modeling slot filling, intent detection and domain classification, implicitly learning the domain-shared and task-shared information.
Such methods can implicitly extract the shared features but fail to effectively capture domain-speciﬁc features.

\textit{Explicit domain knowledge transfer} approaches denote that models used shared-private framework including a shared module to capture domain-shared feature and a private module for each domain, which has the advantage of explicitly differentiating shared and private knowledge. 
\newcite{kim-etal-2017-domain} used attention mechanism to learn a weighted combinations from the feedback of the expert models on different domains.
\newcite{liu2017multidomain} also used a shared LSTM to capture domain-shared knowledge and private LSTM to extract domain-specific feature, combining them for multi-domain slot filling.
\newcite{qin2020multidomain} proposed a model with separate domain- and task-specific parameters, which enables model to capture the task-aware and domain-aware features for multi-domain SLU.

Although there are many methods of cross-domain SLU, main challenges are as follows:
\begin{itemize}
	\item \textbf{Domain Knowledge Transfer:} Transferring knowledge from source domain to target domain is non-trivial.
	In addition, how to conduct domain knowledge transfer at a fine-grained level for \textit{sentence-level} intent detection and \textit{token-level} slot filling is difficult. 
	
	\item \textbf{Zero-shot Setting:} When the target domain has no training data, how to transfer knowledge from source domain data to target domain is a challenge.

\end{itemize}
\subsection{Cross-Lingual SLU}\label{sec:cross-lingual}
Cross-lingual SLU means that an SLU system trained on English can be directly applied to other low-resource languages, which has attracted more and more attention.

In recent years, 
\newcite{schuster-etal-2019-cross-lingual} released a multi-lingual SLU dataset\footnote{https://fb.me/multilingual\_task\_oriented\_data} contains English, Spanish and Thai to facilitate the cross-lingual SLU direction.
\newcite{liu2020attention} proposed an attention-Informed mixed-language training to align representation between source language and target language.
\newcite{xu-etal-2020-end} proposed  a novel end-to-end model
that learns to align and predict target slot labels
jointly for cross-lingual transfer. Besides, they introduced
MultiATIS++, a new multilingual NLU corpus to the community.
\newcite{qin2020codsa-ml} proposed a multilingual code-switching augmentation framework to fine-tune mBERT for aligning representations from source and multiple target languages, which achieved promising performance.
 
Cross-lingual SLU has many interesting problems to focus on:
 \begin{itemize}
 	\item \textbf{Alignment between Source and Target Language:} Aligning intent and slot representations from source and target languages is a core challenge for cross-lingual SLU.
 	
 	\item \textbf{Generalizability:} Since new languages emerge frequently, training a generalized SLU model that can applied to all languages with considerable performance deserves to be explored.
 \end{itemize}

\subsection{Low-resource SLU}\label{sec:low-resource}
The remarkable progress on SLU heavily relies on a considerable amount of labeled training data, which fails to work in low-resources setting where few or zero shot data can be accessed to.
In this section, we will discuss the trends and progress on low-resource SLU, including \textit{Few-shot SLU}, \textit{Zero-shot SLU}, and \textit{Unsupervised SLU}.
\paragraph{\textit{Few-shot SLU:}}\label{sec:few-shot}
In some cases, a slot or intent only has fewer instances, which makes the traditional supervised SLU models powerless.
To alleviate this issue, \textit{few-shot SLU} is appealing in this scenario since it is able to adapt to new application quickly with only very few examples.

Recently, some work has been proposed for investigating this direction.
\newcite{hou-etal-2020-shot} proposed a few-shot CRF model with collapsed dependency transfer mechanism for few-shot slot tagging.
Besides, \newcite{hou2020few} have began to explore the  few-shot multi-intent detection.
\paragraph{\textit{Zero-shot SLU:}}\label{sec:zero-shot}
Face the rapid changing of applications, situations like no target training data may happen in a brand new application. 
Many zero shot methods offer a way of solving it by discovering commonalities between slots.
\newcite{bapna2017towards} proposed a method utilized slot description which carries information of the slots to obtain and transfer concept through different applications, empowering the models' zero-shot ability. 
\newcite{liu-etal-2020-coach} followed similar architecture, which trains the model on awareness of slot descriptions.
\newcite{shah-etal-2019-robust} countered the problems of misaligned overlapping schemas by adding slot examples values along with descriptions during training.

\paragraph{\textit{Unsupervised SLU:}}\label{sec:unsupervised}
In recent years, unsupervised method has been proposed to automatically extract slot-value pairs, which is a promising direction to free model from heavy human annotation.
To this end, \newcite{min2020dialogue} proposed a new task named \textit{dialogue state induction}, which is to automatically identify dialogue state slot-value pairs.

 Low-resource SLU has attracted more and more attention and the main challenges are as follows:
\begin{itemize}
	\item \textbf{Interaction on low-resource setting:} How to make full use of connection between intents and slots in the low-resource setting is still under-explored. 
	
	\item \textbf{Lack of Benchmark:} There is lacking of public benchmarks in low-resource setting, which may impede the progress.
\end{itemize}

\section{Related Work}
\label{related-work}
\newcite{tur2011spoken} gave a summary of the SLU ﬁeld at the advent of the neural era (2011).
\newcite{chen2017survey} 
provided a survey on task-oriented dialog systems, covering a small overall of state-of-the-art SLU models.
\newcite{louvan-magnini-2020-recent} provided a good survey of intent detection and slot filling methods up to 2019.

Compared with their work, the main differences are summarized as follows: (1) In this survey, we introduce a new taxonomy of the technical architecture of SLU and we conduct a comprehensive review of the origin and the development of SLU;
(2) We discuss and analyze the limitation of existing SLU and shed light on some new trends and discuss new frontiers in this research ﬁeld.
(3) We provide an open-sourced website for SLU researchers, hoping to facilitate the SLU field.
We hope that this survey can shed a light on future research in SLU community.
\section{Conclusion}
\label{conclusion}
This article presented a comprehensive survey on the progress of spoken language understanding. Based on a thorough analysis of recent work, we presented a new taxonomy of SLU from diﬀerent modeling perspectives.
In addition,
considering the limitations of recent SLU system, we shed light on some new trends and discuss new frontiers in this research ﬁeld.
Finally, we made the first attempt to provide an open-sourced website including SLU datasets, papers, baseline projects and leaderboards for SLU researchers, hoping to facilitate the SLU community.

\bibliographystyle{named}
\bibliography{ijcai21}

\begin{thebibliography}{}

\bibitem[\protect\citeauthoryear{Bapna \bgroup \em et al.\egroup
  }{2017a}]{bapna-etal-2017-sequential}
Ankur Bapna, Gokhan T{\"u}r, Dilek Hakkani-T{\"u}r, and Larry Heck.
\newblock Sequential dialogue context modeling for spoken language
  understanding.
\newblock In {\em Proc. of {SIG}dial}, 2017.

\bibitem[\protect\citeauthoryear{Bapna \bgroup \em et al.\egroup
  }{2017b}]{bapna2017towards}
Ankur Bapna, Gokhan T{\"u}r, Dilek Hakkani-T{\"u}r, and Larry Heck.
\newblock Towards zero-shot frame semantic parsing for domain scaling.
\newblock In {\em Interspeech}, 2017.

\bibitem[\protect\citeauthoryear{Castellucci \bgroup \em et al.\egroup
  }{2019}]{castellucci2019multilingual}
Giuseppe Castellucci, Valentina Bellomaria, Andrea Favalli, and Raniero
  Romagnoli.
\newblock Multi-lingual intent detection and slot filling in a joint bert-based
  model.
\newblock {\em arXiv preprint arXiv:1907.02884}, 2019.

\bibitem[\protect\citeauthoryear{Chen \bgroup \em et al.\egroup
  }{2016}]{chen2016end-to-end}
Yun-Nung~Vivian Chen, Dilek Hakkani-Tür, Gokhan Tur, Jianfeng Gao, and
  Li~Deng.
\newblock End-to-end memory networks with knowledge carryover for multi-turn
  spoken language understanding.
\newblock In {\em Interspeech}, 2016.

\bibitem[\protect\citeauthoryear{Chen \bgroup \em et al.\egroup
  }{2017}]{chen2017survey}
Hongshen Chen, Xiaorui Liu, Dawei Yin, and Jiliang Tang.
\newblock A survey on dialogue systems: Recent advances and new frontiers.
\newblock {\em ACM SIGkdd Explorations}, 2017.

\bibitem[\protect\citeauthoryear{Chen \bgroup \em et al.\egroup
  }{2019}]{chen2019bert}
Qian Chen, Zhu Zhuo, and Wen Wang.
\newblock Bert for joint intent classification and slot filling.
\newblock {\em arXiv preprint arXiv:1902.10909}, 2019.

\bibitem[\protect\citeauthoryear{Coucke \bgroup \em et al.\egroup
  }{2018}]{coucke2018snips}
Alice Coucke, Alaa Saade, Adrien Ball, Th{\'e}odore Bluche, Alexandre Caulier,
  David Leroy, Cl{\'e}ment Doumouro, Thibault Gisselbrecht, Francesco
  Caltagirone, Thibaut Lavril, et~al.
\newblock Snips voice platform: an embedded spoken language understanding
  system for private-by-design voice interfaces.
\newblock {\em arXiv preprint arXiv:1805.10190}, 2018.

\bibitem[\protect\citeauthoryear{Devlin \bgroup \em et al.\egroup
  }{2019}]{devlin-etal-2019-bert}
Jacob Devlin, Ming-Wei Chang, Kenton Lee, and Kristina Toutanova.
\newblock {BERT}: Pre-training of deep bidirectional transformers for language
  understanding.
\newblock In {\em Proc. of NAACL}, 2019.

\bibitem[\protect\citeauthoryear{E \bgroup \em et al.\egroup
  }{2019}]{e-etal-2019-novel}
Haihong E, Peiqing Niu, Zhongfu Chen, and Meina Song.
\newblock A novel bi-directional interrelated model for joint intent detection
  and slot filling.
\newblock In {\em Proc. of ACL}, 2019.

\bibitem[\protect\citeauthoryear{Gangadharaiah and
  Narayanaswamy}{2019}]{gangadharaiah-narayanaswamy-2019-joint}
Rashmi Gangadharaiah and Balakrishnan Narayanaswamy.
\newblock Joint multiple intent detection and slot labeling for goal-oriented
  dialog.
\newblock In {\em Proc. of NAACL}, 2019.

\bibitem[\protect\citeauthoryear{Goo \bgroup \em et al.\egroup
  }{2018}]{goo-etal-2018-slot}
Chih-Wen Goo, Guang Gao, Yun-Kai Hsu, Chih-Li Huo, Tsung-Chieh Chen, Keng-Wei
  Hsu, and Yun-Nung Chen.
\newblock Slot-gated modeling for joint slot filling and intent prediction.
\newblock In {\em Proc. of NAACL}, 2018.

\bibitem[\protect\citeauthoryear{Hakkani-T{\"u}r \bgroup \em et al.\egroup
  }{2016}]{hakkani2016multi}
Dilek Hakkani-T{\"u}r, G{\"o}khan T{\"u}r, Asli Celikyilmaz, Yun-Nung Chen,
  Jianfeng Gao, Li~Deng, and Ye-Yi Wang.
\newblock Multi-domain joint semantic frame parsing using bi-directional
  rnn-lstm.
\newblock In {\em Interspeech}, 2016.

\bibitem[\protect\citeauthoryear{Hemphill \bgroup \em et al.\egroup
  }{1990}]{hemphill-etal-1990-atis}
Charles~T. Hemphill, John~J. Godfrey, and George~R. Doddington.
\newblock The {ATIS} spoken language systems pilot corpus.
\newblock In {\em Speech and Natural Language: Proc. of a Workshop Held at
  Hidden Valley, {P}ennsylvania, June 24-27,1990}, 1990.

\bibitem[\protect\citeauthoryear{Hochreiter and
  Schmidhuber}{1997}]{hochreiter1997long}
Sepp Hochreiter and J{\"u}rgen Schmidhuber.
\newblock Long short-term memory.
\newblock {\em Neural computation}, 1997.

\bibitem[\protect\citeauthoryear{Hou \bgroup \em et al.\egroup
  }{2020}]{hou-etal-2020-shot}
Yutai Hou, Wanxiang Che, Yongkui Lai, Zhihan Zhou, Yijia Liu, Han Liu, and Ting
  Liu.
\newblock Few-shot slot tagging with collapsed dependency transfer and
  label-enhanced task-adaptive projection network.
\newblock In {\em Proc. of ACL}, 2020.

\bibitem[\protect\citeauthoryear{Hou \bgroup \em et al.\egroup
  }{2021}]{hou2020few}
Yutai Hou, Yongkui Lai, Yushan Wu, Wanxiang Che, and Ting Liu.
\newblock Few-shot learning for multi-label intent detection.
\newblock In {\em Proc. of AAAI}, 2021.

\bibitem[\protect\citeauthoryear{Kim \bgroup \em et al.\egroup
  }{2017a}]{kim2017onenet}
Young-Bum Kim, Sungjin Lee, and Karl Stratos.
\newblock Onenet: Joint domain, intent, slot prediction for spoken language
  understanding.
\newblock In {\em ASRU}, 2017.

\bibitem[\protect\citeauthoryear{Kim \bgroup \em et al.\egroup
  }{2017b}]{kim-etal-2017-domain}
Young-Bum Kim, Karl Stratos, and Dongchan Kim.
\newblock Domain attention with an ensemble of experts.
\newblock In {\em Proc. of ACL}, 2017.

\bibitem[\protect\citeauthoryear{Kurata \bgroup \em et al.\egroup
  }{2016}]{kurata-etal-2016-leveraging}
Gakuto Kurata, Bing Xiang, Bowen Zhou, and Mo~Yu.
\newblock Leveraging sentence-level information with encoder {LSTM} for
  semantic slot filling.
\newblock In {\em Proc. of EMNLP}, 2016.

\bibitem[\protect\citeauthoryear{LeCun \bgroup \em et al.\egroup
  }{1998}]{lecun1998gradient}
Yann LeCun, L{\'e}on Bottou, Yoshua Bengio, and Patrick Haffner.
\newblock Gradient-based learning applied to document recognition.
\newblock {\em Proc. of the IEEE}, 1998.

\bibitem[\protect\citeauthoryear{Li \bgroup \em et al.\egroup
  }{2018}]{li-etal-2018-self}
Changliang Li, Liang Li, and Ji~Qi.
\newblock A self-attentive model with gate mechanism for spoken language
  understanding.
\newblock In {\em Proc. of EMNLP}, 2018.

\bibitem[\protect\citeauthoryear{Liu and Lane}{2015}]{liu2015recurrent}
Bing Liu and Ian Lane.
\newblock Recurrent neural network structured output prediction for spoken
  language understanding.
\newblock In {\em Proc. of NIPS}, 2015.

\bibitem[\protect\citeauthoryear{Liu and Lane}{2016a}]{liu2016attention}
Bing Liu and Ian Lane.
\newblock Attention-based recurrent neural network models for joint intent
  detection and slot filling.
\newblock In {\em Interspeech}, 2016.

\bibitem[\protect\citeauthoryear{Liu and Lane}{2016b}]{liu-lane-2016-joint}
Bing Liu and Ian Lane.
\newblock Joint online spoken language understanding and language modeling with
  recurrent neural networks.
\newblock In {\em Proc. of {SIG}dial}, 2016.

\bibitem[\protect\citeauthoryear{Liu and Lane}{2017}]{liu2017multidomain}
Bing Liu and Ian Lane.
\newblock Multi-domain adversarial learning for slot filling in spoken language
  understanding.
\newblock In {\em NIPS Workshop}, 2017.

\bibitem[\protect\citeauthoryear{Liu \bgroup \em et al.\egroup
  }{2019}]{liu-etal-2019-cm}
Yijin Liu, Fandong Meng, Jinchao Zhang, Jie Zhou, Yufeng Chen, and Jinan Xu.
\newblock {CM}-net: A novel collaborative memory network for spoken language
  understanding.
\newblock In {\em Proc. of EMNLP-IJCNLP}, 2019.

\bibitem[\protect\citeauthoryear{Liu \bgroup \em et al.\egroup
  }{2020a}]{liu2020attention}
Zihan Liu, Genta~Indra Winata, Zhaojiang Lin, Peng Xu, and Pascale Fung.
\newblock Attention-informed mixed-language training for zero-shot
  cross-lingual task-oriented dialogue systems.
\newblock In {\em Proc. of AAAI}, 2020.

\bibitem[\protect\citeauthoryear{Liu \bgroup \em et al.\egroup
  }{2020b}]{liu-etal-2020-coach}
Zihan Liu, Genta~Indra Winata, Peng Xu, and Pascale Fung.
\newblock {C}oach: A coarse-to-fine approach for cross-domain slot filling.
\newblock In {\em Proc. of ACL}, 2020.

\bibitem[\protect\citeauthoryear{Louvan and
  Magnini}{2020}]{louvan-magnini-2020-recent}
Samuel Louvan and Bernardo Magnini.
\newblock Recent neural methods on slot filling and intent classification for
  task-oriented dialogue systems: A survey.
\newblock In {\em Proc. of COLING}, 2020.

\bibitem[\protect\citeauthoryear{Mesnil \bgroup \em et al.\egroup
  }{2013}]{mesnil2013investigation}
Gr{\'e}goire Mesnil, Xiaodong He, Li~Deng, and Yoshua Bengio.
\newblock Investigation of recurrent-neural-network architectures and learning
  methods for spoken language understanding.
\newblock In {\em Interspeech}, 2013.

\bibitem[\protect\citeauthoryear{Mesnil \bgroup \em et al.\egroup
  }{2014}]{mesnil2014using}
Gr{\'e}goire Mesnil, Yann Dauphin, Kaisheng Yao, Yoshua Bengio, Li~Deng, Dilek
  Hakkani-Tur, Xiaodong He, Larry Heck, Gokhan Tur, Dong Yu, et~al.
\newblock Using recurrent neural networks for slot filling in spoken language
  understanding.
\newblock {\em TASLP}, 2014.

\bibitem[\protect\citeauthoryear{Min \bgroup \em et al.\egroup
  }{2020}]{min2020dialogue}
Qingkai Min, Libo Qin, Zhiyang Teng, Xiao Liu, and Yue Zhang.
\newblock Dialogue state induction using neural latent variable models.
\newblock In {\em Proc. of IJCAI}, 2020.

\bibitem[\protect\citeauthoryear{Qin \bgroup \em et al.\egroup
  }{2019}]{qin-etal-2019-stack}
Libo Qin, Wanxiang Che, Yangming Li, Haoyang Wen, and Ting Liu.
\newblock A stack-propagation framework with token-level intent detection for
  spoken language understanding.
\newblock In {\em Proc. of EMNLP-IJCNLP}, 2019.

\bibitem[\protect\citeauthoryear{Qin \bgroup \em et al.\egroup
  }{2020a}]{qin2020codsa-ml}
Libo Qin, Minheng Ni, Yue Zhang, and Wanxiang Che.
\newblock Cosda-ml: Multi-lingual code-switching data augmentation for
  zero-shot cross-lingual nlp.
\newblock In {\em Proc. of IJCAI}, 2020.

\bibitem[\protect\citeauthoryear{Qin \bgroup \em et al.\egroup
  }{2020b}]{qin2020multidomain}
Libo Qin, Minheng Ni, Yue Zhang, Wanxiang Che, Yangming Li, and Ting Liu.
\newblock Multi-domain spoken language understanding using domain-and
  task-aware parameterization.
\newblock {\em arXiv preprint arXiv:2004.14871}, 2020.

\bibitem[\protect\citeauthoryear{Qin \bgroup \em et al.\egroup
  }{2020c}]{qin-etal-2020-agif}
Libo Qin, Xiao Xu, Wanxiang Che, and Ting Liu.
\newblock {AGIF}: An adaptive graph-interactive framework for joint multiple
  intent detection and slot filling.
\newblock In {\em EMNLP Findings}, 2020.

\bibitem[\protect\citeauthoryear{{Qin} \bgroup \em et al.\egroup
  }{2021a}]{9330801}
L.~{Qin}, W.~{Che}, M.~{Ni}, Y.~{Li}, and T.~{Liu}.
\newblock Knowing where to leverage: Context-aware graph convolution network
  with an adaptive fusion layer for contextual spoken language understanding.
\newblock {\em TASLP}, 2021.

\bibitem[\protect\citeauthoryear{Qin \bgroup \em et al.\egroup
  }{2021b}]{qin2020cointeractive}
Libo Qin, Tailu Liu, Wanxiang Che, Bingbing Kang, Sendong Zhao, and Ting Liu.
\newblock A co-interactive transformer for joint slot filling and intent
  detection.
\newblock In {\em ICASSP}, 2021.

\bibitem[\protect\citeauthoryear{Ravuri and
  Stolcke}{2015}]{ravuri2015recurrent}
Suman Ravuri and Andreas Stolcke.
\newblock Recurrent neural network and lstm models for lexical utterance
  classification.
\newblock In {\em Interspeech}, 2015.

\bibitem[\protect\citeauthoryear{Raymond and
  Riccardi}{2007}]{raymond2007generative}
Christian Raymond and Giuseppe Riccardi.
\newblock Generative and discriminative algorithms for spoken language
  understanding.
\newblock In {\em Interspeech}, 2007.

\bibitem[\protect\citeauthoryear{Schuster \bgroup \em et al.\egroup
  }{2019}]{schuster-etal-2019-cross-lingual}
Sebastian Schuster, Sonal Gupta, Rushin Shah, and Mike Lewis.
\newblock Cross-lingual transfer learning for multilingual task oriented
  dialog.
\newblock In {\em Proc. of NAACL}, 2019.

\bibitem[\protect\citeauthoryear{Shah \bgroup \em et al.\egroup
  }{2019}]{shah-etal-2019-robust}
Darsh Shah, Raghav Gupta, Amir Fayazi, and Dilek Hakkani-Tur.
\newblock Robust zero-shot cross-domain slot filling with example values.
\newblock In {\em Proc. of ACL}, 2019.

\bibitem[\protect\citeauthoryear{Su \bgroup \em et al.\egroup
  }{2018}]{su-etal-2018-time}
Shang-Yu Su, Pei-Chieh Yuan, and Yun-Nung Chen.
\newblock How time matters: Learning time-decay attention for contextual spoken
  language understanding in dialogues.
\newblock In {\em Proc. of NAACL}, 2018.

\bibitem[\protect\citeauthoryear{Teng \bgroup \em et al.\egroup
  }{2021}]{teng2020injecting}
Dechuang Teng, Libo Qin, Wanxiang Che, Sendong Zhao, and Ting Liu.
\newblock Injecting word information with multi-level word adapter for chinese
  spoken language understanding.
\newblock In {\em ICASSP}, 2021.

\bibitem[\protect\citeauthoryear{Tjong Kim~Sang and
  De~Meulder}{2003}]{tjong-kim-sang-de-meulder-2003-introduction}
Erik~F. Tjong Kim~Sang and Fien De~Meulder.
\newblock Introduction to the {C}o{NLL}-2003 shared task: Language-independent
  named entity recognition.
\newblock In {\em Proc. of HLT-NAACL}, 2003.

\bibitem[\protect\citeauthoryear{Tur and De~Mori}{2011}]{tur2011spoken}
Gokhan Tur and Renato De~Mori.
\newblock {\em Spoken language understanding: Systems for extracting semantic
  information from speech}.
\newblock 2011.

\bibitem[\protect\citeauthoryear{{Vu} \bgroup \em et al.\egroup
  }{2016}]{Vu2016bi-directional}
N.~T. {Vu}, P.~{Gupta}, H.~{Adel}, and H.~{Schütze}.
\newblock Bi-directional recurrent neural network with ranking loss for spoken
  language understanding.
\newblock In {\em ICASSP}, 2016.

\bibitem[\protect\citeauthoryear{Wang \bgroup \em et al.\egroup
  }{2018}]{wang-etal-2018-bi}
Yu~Wang, Yilin Shen, and Hongxia Jin.
\newblock A bi-model based {RNN} semantic frame parsing model for intent
  detection and slot filling.
\newblock In {\em Proc. of NAACL}, 2018.

\bibitem[\protect\citeauthoryear{Wang \bgroup \em et al.\egroup
  }{2020}]{wang2020from}
Yuxuan Wang, Yutai Hou, Wanxiang Che, and Ting Liu.
\newblock From static to dynamic word representations: a survey.
\newblock {\em IJMLC}, 2020.

\bibitem[\protect\citeauthoryear{Xu and Sarikaya}{2013}]{xu2013convolutional}
Puyang Xu and Ruhi Sarikaya.
\newblock Convolutional neural network based triangular crf for joint intent
  detection and slot filling.
\newblock In {\em ASRU}, 2013.

\bibitem[\protect\citeauthoryear{Xu \bgroup \em et al.\egroup
  }{2020}]{xu-etal-2020-end}
Weijia Xu, Batool Haider, and Saab Mansour.
\newblock End-to-end slot alignment and recognition for cross-lingual {NLU}.
\newblock In {\em Proc. of EMNLP}, 2020.

\bibitem[\protect\citeauthoryear{Yao \bgroup \em et al.\egroup
  }{2013}]{yao2013recurrent}
Kaisheng Yao, Geoffrey Zweig, Mei-Yuh Hwang, Yangyang Shi, and Dong Yu.
\newblock Recurrent neural networks for language understanding.
\newblock In {\em Interspeech}, 2013.

\bibitem[\protect\citeauthoryear{Yao \bgroup \em et al.\egroup
  }{2014a}]{yao2014spoken}
Kaisheng Yao, Baolin Peng, Yu~Zhang, Dong Yu, Geoffrey Zweig, and Yangyang Shi.
\newblock Spoken language understanding using long short-term memory neural
  networks.
\newblock In {\em SLT}, 2014.

\bibitem[\protect\citeauthoryear{Yao \bgroup \em et al.\egroup
  }{2014b}]{yao2014recurrent}
Kaisheng Yao, Baolin Peng, Geoffrey Zweig, Dong Yu, Xiaolong Li, and Feng Gao.
\newblock Recurrent conditional random field for language understanding.
\newblock In {\em ICASSP}, 2014.

\bibitem[\protect\citeauthoryear{Zhang and Wang}{2016}]{zhang2016joint}
Xiaodong Zhang and Houfeng Wang.
\newblock A joint model of intent determination and slot filling for spoken
  language understanding.
\newblock In {\em Proc. of IJCAI}, 2016.

\bibitem[\protect\citeauthoryear{Zhang \bgroup \em et al.\egroup
  }{2019}]{zhang-etal-2019-joint}
Chenwei Zhang, Yaliang Li, Nan Du, Wei Fan, and Philip Yu.
\newblock Joint slot filling and intent detection via capsule neural networks.
\newblock In {\em Proc. of ACL}, 2019.

\bibitem[\protect\citeauthoryear{Zhang \bgroup \em et al.\egroup
  }{2020}]{zhang2020graph}
Linhao Zhang, Dehong Ma, Xiaodong Zhang, Xiaohui Yan, and Houfeng Wang.
\newblock Graph lstm with context-gated mechanism for spoken language
  understanding.
\newblock In {\em Proc. of AAAI}, 2020.

\end{thebibliography}

\end{document}